\documentclass[11pt]{article}
\usepackage[utf8]{inputenc}
\usepackage{jmlr2e}
\usepackage{times}
\usepackage{url}
\usepackage{booktabs}
\usepackage{amsfonts}
\usepackage{nicefrac}
\usepackage{microtype}
\usepackage{bm}
\usepackage{graphicx}
\usepackage{amsmath, empheq, calc}
\usepackage{amsfonts}
\usepackage{pbox}
\usepackage{array}
\usepackage{stmaryrd}
\usepackage{wrapfig}
\usepackage{subcaption}
\usepackage{natbib}
\usepackage{stmaryrd}
\usepackage{multirow}
\usepackage{todonotes}
\usepackage{tikz}
\usepackage{enumitem}
\usepackage{lipsum,multicol}
\usepackage{algorithm}
\usepackage{algorithmic}
\usepackage[toc]{appendix}
\usepackage{color}

\newtheorem{thm}{Theorem}

\usepackage{ifthen}
\def\layersep{1.5}
\def\innersep{.7}
\usetikzlibrary{arrows,calc,positioning,shapes,
                decorations.pathreplacing, shapes.misc, backgrounds}
\tikzset{
    >=stealth',
    punkt/.style={
           circle,
           draw=black, thick,
           minimum height=1.75em,
           inner sep=0pt,
           text centered},
    pil/.style={
           ->,
           thick},
    str/.style={
           -,
           thick},
treenode/.style = {shape=rectangle, rounded corners,
                     draw, align=center,
                     top color=white, bottom color=blue!20},
  root/.style     = {treenode, font=\Large, bottom color=red!30},
  env/.style      = {treenode, font=\ttfamily\normalsize},
  dummy/.style    = {circle,draw},
  cir/.style={
    rectangle,
    rounded corners,
    draw=black, thick,
    minimum width=1.5em,
    text centered},
  arr/.style={
    ->,
    thick,
    shorten <=2pt,
    shorten >=2pt,},
  layer/.style={
    draw=black,
    fill=white,
    thick,
    rectangle,
    rounded corners,
    minimum width=1.5em,
    minimum height=12em
  },
  slayer/.style={
    draw=black,
    fill=white,
    thick,
    rectangle,
    rounded corners,
    minimum width=1.5em,
    minimum height=3em
  },
  brace/.style={
    thick,
    decoration={
    brace,
    mirror,
    raise=2.5cm
    },
    decorate
}}

\newcommand{\gen}[3]{
    \foreach \name / \y in {1,...,4}
       {\ifthenelse{\y<#1}{\node[proba, pin=left: $X_{\y}$] (In-\name) at (-1+#2,\innersep*-\y+#3) {$ a_{\y#1}$};
            }{\ifthenelse{\y>#1}{\pgfmathsetmacro\x{int(\name - 1)} ; \node[proba, pin=left: $X_{\y}$] (In-\x) at (-1+ #2,\innersep*-\y+\innersep+ #3) {$ a_{\y#1}$};
         }{}
       }}
    \node[proba, pin=left: $E_{#1}$] (In-4) at (-1+#2,\innersep*-4+#3) {$1$};
    \foreach \name / \y in {1,...,4}
        \node[input neuron] (I-\name) at (#2,\innersep*-\y+#3) {};
    \pgfmathsetmacro\z{#2+\layersep*0.14};
    \foreach \name / \y in {1,...,5}
              \path[yshift=0.5cm]
            node[hidden neuron] (H-\name) at (\z-.35, \innersep*-\y+#3-.15) {};
 
    \foreach \name / \y in {1,...,5}
        \path[xshift=1cm,yshift=0.5cm]
            node[hidden neuron] (Hi-\name) at (\z,\innersep*-\y+#3-.15) {};

    \node[output neuron,pin={[pin edge={->}]right:$\hat{X_{#1}}$}, right of=H-3, xshift=1.15cm] (O) {};

    \foreach \source in {1,...,4}
        \foreach \dest in {1,...,5}
            \path (I-\source) edge (H-\dest);
    \foreach \dest in {1,...,4}
        \path (In-\dest) edge (I-\dest);
    
    \foreach \dest in {1,...,5}
        \node (q-\dest) at (\z+.35, 1*\innersep*-\dest+#3+.3) {$\cdots$};
    \foreach \source in {1,...,5}
        \path (Hi-\source) edge (O);
    \draw[brace, decoration={raise=1.5cm}] (In-1.north) -- (In-3.south);
    \node[left=1.4cm of In-2, yshift=-0.cm] (inlabel) {${\bf X}_{- #1}$};

}

\newcommand{\blfootnote}[1]{
    \renewcommand{\thefootnote}{}
    \footnotetext{\hspace{-16.5pt}\scriptsize#1}
    \renewcommand{\thefootnote}{\arabic{footnote}}
}

\newcommand{\Pahat}[1]{\text{Pa}({#1}; \widehat{\mathcal{G}})}
\newcommand{\Pahatbar}[1]{\overline{\text{Pa}}({#1}; \widehat{\mathcal{G}})}

\makeatletter
\newcommand*{\indep}{%
  \mathbin{%
    \mathpalette{\@indep}{}%
  }%
}
\newcommand*{\nindep}{%
  \mathbin{
    \mathpalette{\@indep}{\not}
  }%
}
\def\layersep{.38cm}

\newcommand*{\@indep}[2]{%
  \sbox0{$#1\perp\m@th$}
  \sbox2{$#1=$}
  \sbox4{$#1\vcenter{}$}
  \rlap{\copy0}
  \dimen@=\dimexpr\ht2-\ht4-.2pt\relax
  \kern\dimen@
  {#2}%
  \kern\dimen@
  \copy0 
} 
\makeatother

\begin{document}

\title{Structural Agnostic Modeling: Adversarial Learning of\smallskip\\ Causal Graphs}



\author{\name Diviyan Kalainathan$^*$ \email diviyan@fentech.ai\\
\addr Fentech\\
20 Rue Raymond Aron, Paris, France
\AND
\name Olivier Goudet$^*$ \email olivier.goudet@univ-angers.fr\\
\addr LERIA, Université d'Angers\\
\addr 2 boulevard Lavoisier, 49045 Angers, France
\AND
\name Isabelle Guyon \email guyon@chalearn.org\\
\addr TAU, LRI, INRIA, CNRS, Université Paris-Saclay\\
660 Rue Noetzlin, Gif-Sur-Yvette, France
\AND
\name David Lopez-Paz \email dlp@fb.com\\
\addr Facebook AI Research \\
6 Rue Ménars, 75002 Paris
\AND
\name Mich\`ele Sebag \email sebag@lri.fr\\
\addr TAU, LRI, INRIA, CNRS, Université Paris-Saclay\\
660 Rue Noetzlin, Gif-Sur-Yvette, France
}
\editor{}
\maketitle
\blfootnote{$^*$ Equal contribution. This work was done during Olivier Goudet's post-doc at Univ. Paris-Saclay and Diviyan Kalainathan's PhD at Univ. Paris-Saclay. }
\begin{abstract}
A new causal discovery method, {\em Structural Agnostic Modeling} (SAM), is presented in this paper. Leveraging both conditional independencies and distributional asymmetries in the data, SAM aims to find the underlying causal structure from observational data. 
The approach is based on a game between different players estimating each variable distribution conditionally to the others as a neural net, and an adversary aimed at discriminating the generated data with the original data. \textcolor{blue}{A learning criterion combining distribution estimation, sparsity and acyclicity constraints is used to enforce the  optimization of the graph structure and parameters through stochastic gradient descent. SAM is extensively experimentally validated on synthetic and real data.}
\end{abstract}

\def\Untod{\mbox{$[[1,d]]$}}
\def\Unton{\mbox{$[[1,n]]$}}
\def\HatG{\mbox{$\widehat{\mathcal{G}}$}}
\def\HATG{\mbox{$\widehat{\mathcal{G}}$}}
\def\x{\mbox{\bf x}}
\def\X{\mbox{\bf X}}
\def\a{\mbox{\bf a}}
\def\z{\mbox{\bf z}}
\def\btheta{\mbox{\bf $\theta$}}
\def\qjshort{\mbox{$q_j(x_j | x_{\Pahat{j}}, \btheta_j)$}}
\def\qjshortell{\mbox{$q_j(x_j^{(\ell)} | x_{\Pahat{j}}^{(\ell)}, \btheta_j)$}}
\def\acy{acyclicity}

\section{Problem settings \label{sec:problem}}
As said, this paper focuses on causal discovery, that is, finding the DAG $\cal G$ involved in the Functional Causal Model generating the data (section \ref{sec:FCM}). The SAM approach is based on simultaneously learning $d$ Markov kernels, where the $j$-th Markov kernel $q_j$ expresses the conditional density of $X_j$ given its parents in a candidate graph \HATG\  \citep{janzing2010causal} for $j$ ranging in \Untod.

More precisely, these $d$ learning problems are jointly tackled through optimizing the likelihood of the data according to the conditional distributions $q_j(X_j|X_{\Pahat{j}})$, with $X_{\Pahat{j}}$ denoting the estimated causes of $X_j$, while enforcing the sparsity and \acy\ of the graph \HATG\ defined from all edges $X_k \rightarrow X_j$ for $k$ ranging in $\Pahat{j}$.


\subsection{Markov Kernels as functional causal mechanisms}
Let $D = \{\x^{(1)},\ldots, \x^{(n)}\}$ denote the observational dataset, including $n$ iid samples (with $\x^{(\ell)} = (x^{(\ell)}_1, \ldots, x^{(\ell)}_d)$ for $\ell$ ranging in \Unton), sampled from  the unknown joint distribution $p(\mathbf{X})$ corresponding to the sought FCM.

Each Markov kernel $q_j$ is sought as a functional causal mechanism $\widehat{f_j}$:
\begin{equation}
\hat{X}_j = \hat{f}_j([\mathbf{a_j} \odot \mathbf{X},E_j], \btheta_j).
\label{eq:kernel}
\end{equation}
where
\begin{itemize}
    \item $\a_j = (a_{1,j}, \ldots a_{d,j})$ is a binary vector referred to as $j$-th \emph{structural gate}. Coefficient $a_{i,j}$ is 1 iff variable $X_i$ is used to generate $X_j$, that is, edge $X_i \rightarrow X_j$ belongs to graph $\widehat{\mathcal{G}}$. Otherwise, $a_{i,j}$ is set to 0.  Coefficient $a_{i,i}$ is set to 0 to avoid self-loops.\\
    $\Pahat{j}$, defined as the set of indices $i$ such that $a_{i,j}=1$, corresponds to the set of causes of $X_j$ according to $\hat{f}_j$\,;
    \item $\btheta_j$ is a set of parameters (e.g. neural weights) used to compute $\hat{f}_j$\,;
    \item $E_j$ is a noise variable modelling all non observed causes of $X_j$. 
\end{itemize}
In summary, function $\hat{f}_j$ takes as input all variables $X_k$ such that $a_{j,k}=1$, augmented with the noise variable $E_j$, and it is parameterized by $\btheta_j$.

For each sample $\x = (x_1, \ldots, x_d)$, let $\x_{-j}$ be defined as $(x_1, \ldots x_{j-1},x_{j+1}, \ldots, x_d)$. Model $\hat{f}_j$ thus defines a generative model
of $X_j$ conditionally to its estimated causes, noted $q_j(x_j | \mathbf{x}_{-j}, \a_j, \btheta_j)$, or for simplicity \qjshort, as the set $\Pahat{j}$ is fully characterized by the binary vector $\a_j$.

As all noise variables $E_j$ for $j$ in \Untod\ are independent,  all Markov kernels \qjshort\ are \textit{independent} models, making it possible to learn them all in parallel from the observational dataset $D$. 

\subsection{Learning independent Markov kernels}
Learning \qjshort\ consists of learning  $\widehat{f_j}$ and selecting a (minimal) subset of parents $\Pahat{j}$. The solution $\a_j$ and $\theta_j$ is obtained by minimizing the conditional log-likelihood of the data, given by:

\begin{equation}
    S_j^n(\a_j,\theta_j, D) = -\frac{1}{n} \sum_{\ell = 1}^{n} \text{log}\, \qjshortell
    \label{eq:score_problem}
\end{equation}

Following \citep{brown2012conditional}, each 
conditional log-likelihood term is decomposed into three terms as follows, where $p$ is the data distribution:

\begin{multline}\label{eq:rewritting}
  \text{log}\, \qjshortell =  \text{log} \frac{q_j(x^{(\ell)}_j | x^{(\ell)}_{\Pahat{j}}, \btheta_j)}{p(x^{(\ell)}_j | x^{(\ell)}_{\Pahat{j}})} + \text{log} \frac{p(x^{(\ell)}_j |  x^{(\ell)}_{\Pahat{j}})}{p(x^{(\ell)}_j | \x^{(\ell)}_{-j})} + \text{log} \,{p(x^{(\ell)}_j |  \x^{(\ell)}_{-j})}
\end{multline}

Note that the sum $\frac{1}{n}  \sum_{\ell = 1}^{n} \text{log}\ {p(x^{(\ell)}_j |  \x^{(\ell)}_{-j})}$ converges toward the constant $H(X_j|\X_{-j})$ as $n$ goes to infinity; it is thus discarded in the following.

Let $X_{\Pahatbar{j}}$ denote the complementary set of $X_j$ and its parent nodes in \HATG. Then, after 
\cite{brown2012conditional},  $\frac{1}{n}  \sum_{\ell = 1}^{n}  \text{log} \frac{p(x^{(\ell)}_j | \x^{(\ell)}_{-j})}{p(x^{(\ell)}_j |  x^{(\ell)}_{\Pahat{j}})}$ is equal to the 
empirical conditional mutual information term between $X_j$ and $X_{\Pahatbar{j}}$, conditioned on the parent variables $X_{\Pahat{j}}$:
\begin{equation}\label{condMI}
\hat{I}^n(X_j,X_{\Pahatbar{j}}|X_{\Pahat{j}}) = \frac{1}{n}  \sum_{\ell = 1}^{n} \log \frac{p(x^{(\ell)}_j,x^{(\ell)}_{\Pahatbar{j}}|x^{(\ell)}_{\Pahat{j}})}{p(x^{(\ell)}_j|x^{(\ell)}_{\Pahat{j}})p(x^{(\ell)}_{\Pahatbar{j}}|x^{(\ell)}_{\Pahat{j}})}
\end{equation}

Eventually, the negative conditional log-likelihood score (Eq. (\ref{eq:score_problem})) can be rewritten as:

\begin{equation}
    S_j^n(\a_j,\theta_j, D) = \hat{I}^n(X_j,X_{\Pahatbar{j}}|X_{\Pahat{j}}) + \frac{1}{n}  \sum_{\ell = 1}^{n} \text{log} \frac{p(x^{(\ell)}_j | x^{(\ell)}_{\Pahat{j}})}{q(x^{(\ell)}_j | x^{(\ell)}_{\Pahat{j}}, \btheta_j)}  + cst
    \label{eq:new_score}
\end{equation}

The term $\hat{I}^n(X_j,X_{\Pahatbar{j}}|X_{\Pahat{j}})$ is used to identify the Markov equivalence class of the true $\cal G$, while the
term $\frac{1}{n}  \sum_{\ell = 1}^{n} \text{log} \frac{p(x^{(\ell)}_j | x^{(\ell)}_{\Pahat{j}})}{q(x^{(\ell)}_j | x^{(\ell)}_{\Pahat{j}}, \btheta_j)}$
 is used to disambiguate graphs within the Markov equivalence class of $\cal G$.
Both terms are discussed in the following two subsections.

\subsection{Structural loss \label{sec:structural_loss}}

For each Markov kernel, the minimization of  $\hat{I}^n(X_j,X_{\Pahatbar{j}}|X_{\Pahat{j}})$ (Eq. (\ref{eq:new_score})) corresponds to a feature selection problem, the selection of $X_{\Pahat{j}}$. As shown by \citep{yu2018unified},\footnote{Note that $\hat{I}^n(X_j,X_{\Pahatbar{j}}|X_{\Pahat{j}})$ converges in probability toward $I(X_j,X_{\Pahatbar{j}}|X_{\Pahat{j}})$, the mutual information term between $X_j$ and $X_{\Pahatbar{j}}$, conditioned on the parent variables $X_{\Pahat{j}}$, as $n$ goes to infinity.} the solution of this feature selection problem converges toward 
the Markov Blanket $MB(X_j)$ of $X_j$ in the true causal graph $\mathcal{G}$ in the large sample limit. 
 
This feature selection problem is classically tackled by optimizing the log-likelihood of the data augmented with a regularization term of the form $\lambda_S |\Pahat{j}|$, with $|\Pahat{j}|$ the number of parents of $X_j$ in $\widehat{\mathcal{G}}$ and hyper-parameter $\lambda_S > 0$ controlling the sparsity of selection in the Markov Blanket.\footnote{By construction, $|\Pahat{j}| =\sum_{i=1}^d a_{i,j}$ corresponds to the $L_1$ norm of vector $\a_j$.}

Therefore, without acyclicity constraint, the optimization of the following structural loss enables to identify the \textit{moral graph} associated with the true causal graph $\mathcal{G}$:
\begin{equation}
\mathcal{L}^n_S(\widehat{\mathcal{G}}, D) = \sum_{j=1}^{d}  \hat{I}^n(X_j,X_{\Pahatbar{j}}|X_{\Pahat{j}}) \, + \lambda_S |\widehat{\mathcal{G}}|.
\label{eq:moral}
\end{equation}

A first contribution of the proposed approach is to establish that, searching a DAG minimizing  Eq. (\ref{eq:moral}), leads to identify the Markov equivalence class of $\mathcal{G}$ (CPDAG) in the large sample limit. The intuition is that the \acy\ constraint prevents the children nodes from being selected as parents, hence the spouse nodes do not need be selected either.\footnote{Note that algorithms such as GENIE3 \citep{irrthum2010inferring}, winner of the DREAM4 and DREAM5 challenges, also rely on solving $d$ independent feature selection problems in parallel, but without any \acy\ constraint. They might thus incur some false discovery rate (selecting edges that are not in $\mathcal{G}$). We shall return to this point in section \ref{sec:first_exp}.} 

\begin{thm}[CPDAG identification by structural loss minimization] \label{Prop1}\ \\

\noindent Under  CMA, CFA and CSA assumptions, two results of convergence in probability, hold: \\
i) For every DAG \HATG\ in the equivalence class of $\mathcal{G}$, 
\[ \lim_{n\to\infty} \mathcal{L}^n_S(\widehat{\mathcal{G}}, D) = \mathcal{L}^n_S(\mathcal{G}, D)\]
ii) For every DAG \HATG\ not in the equivalence class of $\mathcal{G}$,  there exists $\lambda_S > 0$ such that:
\[ \lim_{n\to\infty} \mathbb{P}(\mathcal{L}^n_S(\widehat{\mathcal{G}}, D) > \mathcal{L}^n_S(\mathcal{G}, D)) = 1\]
\end{thm}

\begin{proof}
in Appendix  \ref{proofCPDAG}
\end{proof}

Experimental and analytical illustrations of this result on the toy 3-variable skeleton  $A-B-C$ are presented in Appendix \ref{proofCPDAG}.\\

The limitation of the structural loss is that it does not allow one to disambiguate among equivalent DAGs. Typically in the bivariate case, both graphs ($X \rightarrow Y$ and $Y \rightarrow X$) get the same structural loss in the large sample limit ($I(X,Y) + \lambda_S$). We shall see that the parametric loss addresses this limitation.

\subsection{Parametric loss \label{sec:parametric_loss}}

The second term in Eq. (\ref{eq:new_score}), $\frac{1}{n}  \sum_{\ell = 1}^{n} \text{log} \frac{p(x^{(\ell)}_j | x^{(\ell)}_{\Pahat{j}})}{q_j(x^{(\ell)}_j | x^{(\ell)}_{\Pahat{j}}, \btheta_j)}$, measures the ability of $\hat{f}_j$ to fit the conditional distribution of $X_j$ based on its parents $X_\Pahat{j}$. 

Note that in the large sample limit, this term converges towards $\mathbb{E}_p\left[\log \frac{p(x_j|x_{\Pahat{j}})}{ q_j(x_j|x_{\Pahat{j}},\theta_j)} \right]$, and it goes to 0 when considering sufficiently powerful causal mechanisms, irrespective of whether $\widehat{\mathcal{G}} \neq \mathcal{G}$: As shown by \cite{hyvarinen1999nonlinear}, it is always possible to find a function $\hat{f}_j$ such that $X_j \sim  \hat{f}_j(X_{\Pahat{j}},E_j)$, with $E_j \indep X_{\Pahat{j}}$, corresponding to a probabilistic conditional model $q$ such that $q_j(x_j|x_{\Pahat{j}},\theta_j) = p(x_j|x_{\Pahat{j}})$.

In order to support model identification within the Markov equivalence class of the true DAG, a principled approach is to restrict the hypothesis space \citep{hoyer2009nonlinear,zhang2010distinguishing}. In counterpart, such restrictions limit the generality of the approach and may cause practical problems, particularly so when there is no information available about the true generative mechanisms of the data. Therefore, taking inspiration from GPI  pioneering approach \citep{stegle2010probabilistic}, we propose to restrict the capacity of the causal mechanisms $\hat f_j$ through a regularization term. 
Algorithmically, the complexity of the causal mechanisms is controlled through using the Frobenius norm of the parameters in $\hat f_j$ as regularization term \citep{neyshabur2017exploring}, with regularization weight $\lambda_F$. Considering other regularization terms is left for further work. 

Eventually, the parametric loss is defined as the sum of the data fitting terms and the regularization term:
\begin{equation}
\mathcal{L}^n_F(\widehat{\mathcal{G}},\theta, D) = \sum_{j=1}^{d} \left[\frac{1}{n}  \sum_{\ell = 1}^{n} \text{log} \frac{p(x^{(\ell)}_j | x^{(\ell)}_{\Pahat{j}})}{q_j(x^{(\ell)}_j | x^{(\ell)}_{\Pahat{j}}, \theta_j)} ] + \lambda_F\|\theta_j\|_F \right]
\end{equation}

How this parametric loss can disambiguate among the different models in the CPDAG is illustrated in Appendix \ref{expeBivariate}.

\subsection{Discussion}
Eventually, the proposed approach aims to search a DAG $\widehat{\mathcal{G}}$ optimizing a trade-off between 
the data fitting loss, the structural and parametric regularization terms: 
 \begin{equation}
 \begin{array}{ll}
 S^n(\widehat{\mathcal{G}},\theta, D) & := \sum_{j=1}^d  S_j^n(\a_j,\theta_j, D)  + \lambda_S \sum_{j=1}^d \sum_{i=1}^d a_{i,j} +  \lambda_F \sum_{j=1}^d  \|\theta_j\|_F \\
 &=  - \underbrace{\frac{1}{n} \sum_{j=1}^d \sum_{\ell = 1}^{n} {\text{log}\  q_j(x^{(\ell)}_j | x^{(\ell)}_{\Pahat{j}}, \btheta_j)}}_\textrm{data fitting} + \underbrace{  \lambda_S \sum_{j=1}^d \sum_{i=1}^d a_{i,j}  +   \lambda_F \sum_{j=1}^d  \|\theta_j\|_F}_\textrm{model complexity}\\
 &= \underbrace{\sum_{j=1}^{d} \left[ \hat{I}^n(X_j,X_{\Pahatbar{j}}|X_{\Pahat{j}}) \right] + \lambda_S |\widehat{\mathcal{G}}|}_\textrm{structural loss} +  \underbrace{\sum_{j=1}^{d} \left[\frac{1}{n}  \sum_{\ell = 1}^{n} \text{log} \frac{p(x^{(\ell)}_j | x^{(\ell)}_{\Pahat{j}})}{q(x^{(\ell)}_j | x^{(\ell)}_{\Pahat{j}}, \theta_j)} ] + \lambda_F\|\theta_j\|_F \right]}_\textrm{parametric loss}
 \end{array}\label{eq:global_score}
 \end{equation}

As said, the model complexity of the causal mechanisms is decomposed into the {\em structural complexity} (the $L_0$ norm of the structural gates, that is the number of edges in \HATG) and the {\em functional complexity} (the Frobenius norm of the parameters involved in each $\hat{f}_j$). Seen differently, the proposed approach aims to search a DAG that simultaneously minimizes the {\em structural loss} (section \ref{sec:structural_loss}) and the {\em parametric loss} (section \ref{sec:parametric_loss}).

The {\em structural loss} akin category I, II and III approaches \citep{spirtes2000causation,chickering2002optimal} (section \ref{sec:soa})
aims to identify  the Markov equivalence class of the true $\cal G$, while the  {\em parametric loss} akin cause effect pair methods \citep{stegle2010probabilistic},  exploits distribution asymmetries to disambiguate models in the CPDAG equivalence class of $\cal G$.

Note that this approach can accommodate any available prior knowledge about the generative mechanisms of the data, regarding either the type and complexity of the causal mechanisms (e.g. linear or polynomial functions) or the noise distributions (e.g. Gaussian or uniform noise). 

In order to demonstrate the applicability of the approach in the general case (where there exists little or no information about the generative mechanisms of the data), the Structural Agnostic Modelling algorithm 
uses neural networks to model the causal mechanisms $\hat{f_j}$, and relies on adversarial learning to optimize the  data fitting (conditional likelihood) terms. 
Note that the minimisation of Eq. (\ref{eq:global_score}) does not guarantee to obtain a DAG. Therefore, a global acyclicity constraint will be introduced in section \ref{sec:acyclicity_constraint}.  It will serve to couple the learning of the $d$ Markov kernels in parallel. 


\begin{thebibliography}{86}
\providecommand{\natexlab}[1]{#1}
\providecommand{\url}[1]{\texttt{#1}}
\expandafter\ifx\csname urlstyle\endcsname\relax
  \providecommand{\doi}[1]{doi: #1}\else
  \providecommand{\doi}{doi: \begingroup \urlstyle{rm}\Url}\fi

\bibitem[Aliferis et~al.(2003)Aliferis, Tsamardinos, and
  Statnikov]{aliferis2003hiton}
Constantin~F Aliferis, Ioannis Tsamardinos, and Alexander Statnikov.
\newblock Hiton: a novel {Markov} blanket algorithm for optimal variable
  selection.
\newblock In \emph{AMIA annual symposium proceedings}, volume 2003, page~21.
  American Medical Informatics Association, 2003.

\bibitem[Aliferis et~al.(2010)Aliferis, Statnikov, Tsamardinos, Mani, and
  Koutsoukos]{aliferis2010local}
Constantin~F Aliferis, Alexander Statnikov, Ioannis Tsamardinos, Subramani
  Mani, and Xenofon~D Koutsoukos.
\newblock Local causal and {Markov} blanket induction for causal discovery and
  feature selection for classification part {I}: Algorithms and empirical
  evaluation.
\newblock \emph{Journal of Machine Learning Research}, 11\penalty0 (1), 2010.

\bibitem[Aragam and Zhou(2015)]{aragam2015concave}
Bryon Aragam and Qing Zhou.
\newblock Concave penalized estimation of sparse gaussian bayesian networks.
\newblock \emph{Journal of Machine Learning Research}, 16:\penalty0 2273--2328,
  2015.

\bibitem[Aragam et~al.(2017)Aragam, Gu, and Zhou]{aragam2017learning}
Bryon Aragam, Jiaying Gu, and Qing Zhou.
\newblock Learning large-scale bayesian networks with the sparsebn package.
\newblock \emph{arXiv preprint arXiv:1703.04025}, 2017.

\bibitem[Bell and Wang(2000)]{bell2000formalism}
David~A Bell and Hui Wang.
\newblock A formalism for relevance and its application in feature subset
  selection.
\newblock \emph{Machine learning}, 41\penalty0 (2):\penalty0 175--195, 2000.

\bibitem[Bl{\"o}baum et~al.(2018)Bl{\"o}baum, Janzing, Washio, Shimizu, and
  Sch{\"o}lkopf]{blobaum2018cause}
Patrick Bl{\"o}baum, Dominik Janzing, Takashi Washio, Shohei Shimizu, and
  Bernhard Sch{\"o}lkopf.
\newblock Cause-effect inference by comparing regression errors.
\newblock In \emph{International Conference on Artificial Intelligence and
  Statistics}, pages 900--909. PMLR, 2018.

\bibitem[Brown et~al.(2012)Brown, Pocock, Zhao, and
  Luj{\'a}n]{brown2012conditional}
Gavin Brown, Adam Pocock, Ming-Jie Zhao, and Mikel Luj{\'a}n.
\newblock Conditional likelihood maximisation: a unifying framework for
  information theoretic feature selection.
\newblock \emph{Journal of machine learning research}, 13\penalty0
  (Jan):\penalty0 27--66, 2012.

\bibitem[B{\"u}hlmann et~al.(2014)B{\"u}hlmann, Peters, Ernest,
  et~al.]{buhlmann2014cam}
Peter B{\"u}hlmann, Jonas Peters, Jan Ernest, et~al.
\newblock Cam: Causal additive models, high-dimensional order search and
  penalized regression.
\newblock \emph{The Annals of Statistics}, 42\penalty0 (6):\penalty0
  2526--2556, 2014.

\bibitem[Chen et~al.(2007)Chen, Hsiao, Flaschel, and Semmler]{chen2007causal}
Pu~Chen, Chihying Hsiao, Peter Flaschel, and Willi Semmler.
\newblock Causal analysis in economics: Methods and applications.
\newblock 2007.

\bibitem[Chickering(2002)]{chickering2002optimal}
David~Maxwell Chickering.
\newblock Optimal structure identification with greedy search.
\newblock \emph{Journal of Machine Learning Research}, 2002.

\bibitem[Chickering(2013)]{chickering2013transformational}
David~Maxwell Chickering.
\newblock A transformational characterization of equivalent bayesian network
  structures.
\newblock \emph{arXiv preprint arXiv:1302.4938}, 2013.

\bibitem[Colombo and Maathuis(2014)]{colombo2014order}
Diego Colombo and Marloes~H Maathuis.
\newblock Order-independent constraint-based causal structure learning.
\newblock \emph{Journal of Machine Learning Research}, 2014.

\bibitem[Colombo et~al.(2012)Colombo, Maathuis, Kalisch, and
  Richardson]{colombo2012learning}
Diego Colombo, Marloes~H Maathuis, Markus Kalisch, and Thomas~S Richardson.
\newblock Learning high-dimensional directed acyclic graphs with latent and
  selection variables.
\newblock \emph{The Annals of Statistics}, 2012.

\bibitem[Daniusis et~al.(2012)Daniusis, Janzing, Mooij, Zscheischler, Steudel,
  Zhang, and Sch{\"o}lkopf]{daniusis2012inferring}
Povilas Daniusis, Dominik Janzing, Joris Mooij, Jakob Zscheischler, Bastian
  Steudel, Kun Zhang, and Bernhard Sch{\"o}lkopf.
\newblock Inferring deterministic causal relations.
\newblock \emph{arXiv}, 2012.

\bibitem[Doshi-Velez and Kim(2017)]{Velez17}
F.~Doshi-Velez and B.~Kim.
\newblock Towards a rigorous science of interpretable machine learning.
\newblock \emph{arXiv:1702.08608}, 2017.

\bibitem[Faith et~al.(2007)Faith, Hayete, Thaden, Mogno, Wierzbowski, Cottarel,
  Kasif, Collins, and Gardner]{faith2007large}
Jeremiah~J Faith, Boris Hayete, Joshua~T Thaden, Ilaria Mogno, Jamey
  Wierzbowski, Guillaume Cottarel, Simon Kasif, James~J Collins, and Timothy~S
  Gardner.
\newblock Large-scale mapping and validation of escherichia coli
  transcriptional regulation from a compendium of expression profiles.
\newblock \emph{PLoS biol}, 5\penalty0 (1):\penalty0 e8, 2007.

\bibitem[Friedman and Nachman(2000)]{Friedman2000}
Nir Friedman and Iftach Nachman.
\newblock Gaussian process networks.
\newblock In \emph{Proceedings of the Sixteenth Conference on Uncertainty in
  Artificial Intelligence}, UAI'00, page 211–219, San Francisco, CA, USA,
  2000. Morgan Kaufmann Publishers Inc.
\newblock ISBN 1558607099.

\bibitem[Goodfellow et~al.(2014)Goodfellow, Pouget-Abadie, Mirza, Xu,
  Warde-Farley, Ozair, Courville, and Bengio]{goodfellow2014generative}
Ian Goodfellow, Jean Pouget-Abadie, Mehdi Mirza, Bing Xu, David Warde-Farley,
  Sherjil Ozair, Aaron Courville, and Yoshua Bengio.
\newblock Generative adversarial nets.
\newblock \emph{{Advances in Neural Information Processing Systems}}, 2014.

\bibitem[Goudet et~al.(2018)Goudet, Kalainathan, Caillou, Guyon, Lopez-Paz, and
  Sebag]{goudet2018learning}
Olivier Goudet, Diviyan Kalainathan, Philippe Caillou, Isabelle Guyon, David
  Lopez-Paz, and Michele Sebag.
\newblock Learning functional causal models with generative neural networks.
\newblock In \emph{Explainable and Interpretable Models in Computer Vision and
  Machine Learning}, pages 39--80. Springer, 2018.

\bibitem[Gretton et~al.(2005)Gretton, Herbrich, Smola, Bousquet, and
  Sch{\"o}lkopf]{gretton2005kernel}
Arthur Gretton, Ralf Herbrich, Alexander Smola, Olivier Bousquet, and Bernhard
  Sch{\"o}lkopf.
\newblock Kernel methods for measuring independence.
\newblock \emph{Journal of Machine Learning Research}, 2005.

\bibitem[Gretton et~al.(2007)Gretton, Borgwardt, Rasch, Sch{\"o}lkopf, Smola,
  et~al.]{gretton2007kernel}
Arthur Gretton, Karsten~M Borgwardt, Malte Rasch, Bernhard Sch{\"o}lkopf,
  Alexander~J Smola, et~al.
\newblock A kernel method for the two-sample-problem.
\newblock \emph{{Advances in Neural Information Processing Systems}}, 2007.

\bibitem[Guyon et~al.(2019)Guyon, Statnikov, and Bakir~Batu]{guyon2019cause}
Isabelle Guyon, Alexander Statnikov, and Berna Bakir~Batu.
\newblock \emph{Cause Effect Pairs in Machine Learning}.
\newblock Springer International Publishing, 2019.
\newblock ISBN 978-3-030-21809-6.

\bibitem[Haury et~al.(2012)Haury, Mordelet, Vera-Licona, and
  Vert]{haury2012tigress}
Anne-Claire Haury, Fantine Mordelet, Paola Vera-Licona, and Jean-Philippe Vert.
\newblock Tigress: trustful inference of gene regulation using stability
  selection.
\newblock \emph{BMC systems biology}, 6\penalty0 (1):\penalty0 145, 2012.

\bibitem[Hauser and B{\"u}hlmann(2012)]{hauser2012characterization}
Alain Hauser and Peter B{\"u}hlmann.
\newblock Characterization and greedy learning of interventional {Markov}
  equivalence classes of directed acyclic graphs.
\newblock \emph{Journal of Machine Learning Research}, 13\penalty0
  (Aug):\penalty0 2409--2464, 2012.

\bibitem[Hjelm et~al.(2017)Hjelm, Jacob, Che, Trischler, Cho, and
  Bengio]{hjelm2017boundary}
R~Devon Hjelm, Athul~Paul Jacob, Tong Che, Adam Trischler, Kyunghyun Cho, and
  Yoshua Bengio.
\newblock Boundary-seeking generative adversarial networks.
\newblock \emph{arXiv preprint arXiv:1702.08431}, 2017.

\bibitem[Hoyer et~al.(2009)Hoyer, Janzing, Mooij, Peters, and
  Sch{\"o}lkopf]{hoyer2009nonlinear}
Patrik~O Hoyer, Dominik Janzing, Joris~M Mooij, Jonas Peters, and Bernhard
  Sch{\"o}lkopf.
\newblock Nonlinear causal discovery with additive noise models.
\newblock \emph{{Advances in Neural Information Processing Systems}}, 2009.

\bibitem[Hyv{\"a}rinen and Pajunen(1999)]{hyvarinen1999nonlinear}
Aapo Hyv{\"a}rinen and Petteri Pajunen.
\newblock Nonlinear independent component analysis: Existence and uniqueness
  results.
\newblock \emph{Neural Networks}, 12\penalty0 (3):\penalty0 429--439, 1999.

\bibitem[Imbens and Rubin(2015)]{Imbens_book}
Guido~W Imbens and Donald~B Rubin.
\newblock \emph{Causal inference in statistics, social, and biomedical
  sciences}.
\newblock Cambridge University Press, 2015.

\bibitem[Ioffe and Szegedy(2015)]{ioffe2015batch}
Sergey Ioffe and Christian Szegedy.
\newblock Batch normalization: Accelerating deep network training by reducing
  internal covariate shift.
\newblock In \emph{International conference on machine learning}, pages
  448--456, 2015.

\bibitem[Irrthum et~al.(2010)Irrthum, Wehenkel, Geurts,
  et~al.]{irrthum2010inferring}
Alexandre Irrthum, Louis Wehenkel, Pierre Geurts, et~al.
\newblock Inferring regulatory networks from expression data using tree-based
  methods.
\newblock \emph{PloS one}, 5\penalty0 (9):\penalty0 e12776, 2010.

\bibitem[Jang et~al.(2016)Jang, Gu, and Poole]{jang2016categorical}
Eric Jang, Shixiang Gu, and Ben Poole.
\newblock Categorical reparameterization with gumbel-softmax.
\newblock \emph{arXiv preprint arXiv:1611.01144}, 2016.

\bibitem[Janzing and Scholkopf(2010)]{janzing2010causal}
Dominik Janzing and Bernhard Scholkopf.
\newblock Causal inference using the algorithmic {Markov} condition.
\newblock \emph{IEEE Transactions on Information Theory}, 56\penalty0
  (10):\penalty0 5168--5194, 2010.

\bibitem[Janzing and Sch{\"o}lkopf(2018)]{janzing2018detecting}
Dominik Janzing and Bernhard Sch{\"o}lkopf.
\newblock Detecting confounding in multivariate linear models via spectral
  analysis.
\newblock \emph{Journal of Causal Inference}, 6\penalty0 (1), 2018.

\bibitem[Kalisch and B{\"u}hlmann(2007)]{kalisch2007estimating}
Markus Kalisch and Peter B{\"u}hlmann.
\newblock Estimating high-dimensional directed acyclic graphs with the
  pc-algorithm.
\newblock \emph{Journal of Machine Learning Research}, 8\penalty0
  (Mar):\penalty0 613--636, 2007.

\bibitem[Kalisch et~al.(2012)Kalisch, M{\"a}chler, Colombo, Maathuis,
  B{\"u}hlmann, et~al.]{kalisch2012causal}
Markus Kalisch, Martin M{\"a}chler, Diego Colombo, Marloes~H Maathuis, Peter
  B{\"u}hlmann, et~al.
\newblock Causal inference using graphical models with the {R} package pcalg.
\newblock \emph{Journal of Statistical Software}, 2012.

\bibitem[Karras et~al.(2017)Karras, Aila, Laine, and
  Lehtinen]{karras2017progressive}
Tero Karras, Timo Aila, Samuli Laine, and Jaakko Lehtinen.
\newblock Progressive growing of {GANs} for improved quality, stability, and
  variation.
\newblock \emph{arXiv preprint arXiv:1710.10196}, 2017.

\bibitem[{Kingma} and {Ba}(2014)]{2014arXiv1412.6980K}
Durk~P {Kingma} and Jimmy {Ba}.
\newblock {Adam: A Method for Stochastic Optimization}.
\newblock \emph{Int. Conf. on Learning Representations}, 2014.

\bibitem[K{\"u}ffner et~al.(2012)K{\"u}ffner, Petri, Tavakkolkhah, Windhager,
  and Zimmer]{kuffner2012inferring}
Robert K{\"u}ffner, Tobias Petri, Pegah Tavakkolkhah, Lukas Windhager, and Ralf
  Zimmer.
\newblock Inferring gene regulatory networks by anova.
\newblock \emph{Bioinformatics}, 28\penalty0 (10):\penalty0 1376--1382, 2012.

\bibitem[Leray and Gallinari(1999)]{leray1999feature}
Philippe Leray and Patrick Gallinari.
\newblock Feature selection with neural networks.
\newblock \emph{Behaviormetrika}, 26\penalty0 (1):\penalty0 145--166, 1999.

\bibitem[Lopez-Paz and Oquab(2016)]{lopez2016revisiting}
David Lopez-Paz and Maxime Oquab.
\newblock Revisiting classifier two-sample tests.
\newblock \emph{arXiv preprint arXiv:1610.06545}, 2016.

\bibitem[Lopez-Paz et~al.(2015)Lopez-Paz, Muandet, Sch{\"o}lkopf, and
  Tolstikhin]{lopez2015towards}
David Lopez-Paz, Krikamol Muandet, Bernhard Sch{\"o}lkopf, and Ilya~O
  Tolstikhin.
\newblock Towards a learning theory of cause-effect inference.
\newblock \emph{Int. Conf. on Machine Learning}, 2015.

\bibitem[Louizos et~al.(2017)Louizos, Welling, and Kingma]{louizos2017learning}
Christos Louizos, Max Welling, and Diederik~P Kingma.
\newblock Learning sparse neural networks through $ l\_0 $ regularization.
\newblock \emph{arXiv preprint arXiv:1712.01312}, 2017.

\bibitem[Maddison et~al.(2016)Maddison, Mnih, and Teh]{maddison2016concrete}
Chris~J Maddison, Andriy Mnih, and Yee~Whye Teh.
\newblock The concrete distribution: A continuous relaxation of discrete random
  variables.
\newblock \emph{arXiv preprint arXiv:1611.00712}, 2016.

\bibitem[Marbach et~al.(2009)Marbach, Schaffter, Floreano, Prill, and
  Stolovitzky]{marbach2009dream4}
Daniel Marbach, Thomas Schaffter, Dario Floreano, Robert~J Prill, and Gustavo
  Stolovitzky.
\newblock The dream4 in-silico network challenge.
\newblock \emph{Draft, version 0.3}, 2009.

\bibitem[Marbach et~al.(2012)Marbach, Costello, K{\"u}ffner, Vega, Prill,
  Camacho, Allison, Kellis, Collins, and Stolovitzky]{marbach2012wisdom}
Daniel Marbach, James~C Costello, Robert K{\"u}ffner, Nicole~M Vega, Robert~J
  Prill, Diogo~M Camacho, Kyle~R Allison, Manolis Kellis, James~J Collins, and
  Gustavo Stolovitzky.
\newblock Wisdom of crowds for robust gene network inference.
\newblock \emph{Nature methods}, 9\penalty0 (8):\penalty0 796--804, 2012.

\bibitem[Margolin et~al.(2006)Margolin, Nemenman, Basso, Wiggins, Stolovitzky,
  Dalla~Favera, and Califano]{margolin2006aracne}
Adam~A Margolin, Ilya Nemenman, Katia Basso, Chris Wiggins, Gustavo
  Stolovitzky, Riccardo Dalla~Favera, and Andrea Califano.
\newblock Aracne: an algorithm for the reconstruction of gene regulatory
  networks in a mammalian cellular context.
\newblock In \emph{BMC bioinformatics}, volume~7, page~S7. Springer, 2006.

\bibitem[Mendes et~al.(2003)Mendes, Sha, and Ye]{mendes2003artificial}
Pedro Mendes, Wei Sha, and Keying Ye.
\newblock Artificial gene networks for objective comparison of analysis
  algorithms.
\newblock \emph{Bioinformatics}, 19\penalty0 (suppl\_2):\penalty0 ii122--ii129,
  2003.

\bibitem[Mirza and Osindero(2014)]{mirza2014conditional}
Mehdi Mirza and Simon Osindero.
\newblock Conditional generative adversarial nets.
\newblock \emph{arXiv}, 2014.

\bibitem[Mooij et~al.(2010)Mooij, Stegle, Janzing, Zhang, , and
  Sch{\"o}lkopf]{stegle2010probabilistic}
Joris~M Mooij, Oliver Stegle, Dominik Janzing, Kun Zhang, , and Bernhard
  Sch{\"o}lkopf.
\newblock Probabilistic latent variable models for distinguishing between cause
  and effect.
\newblock \emph{{Advances in Neural Information Processing Systems}}, 2010.

\bibitem[Mooij et~al.(2016)Mooij, Peters, Janzing, Zscheischler, and
  Sch{\"o}lkopf]{mooij2016distinguishing}
Joris~M Mooij, Jonas Peters, Dominik Janzing, Jakob Zscheischler, and Bernhard
  Sch{\"o}lkopf.
\newblock Distinguishing cause from effect using observational data: methods
  and benchmarks.
\newblock \emph{Journal of Machine Learning Research}, 2016.

\bibitem[Nandy et~al.(2015)Nandy, Hauser, and Maathuis]{nandy2015high}
Preetam Nandy, Alain Hauser, and Marloes~H Maathuis.
\newblock High-dimensional consistency in score-based and hybrid structure
  learning.
\newblock \emph{arXiv}, 2015.

\bibitem[Neyshabur et~al.(2017)Neyshabur, Bhojanapalli, McAllester, and
  Srebro]{neyshabur2017exploring}
Behnam Neyshabur, Srinadh Bhojanapalli, David McAllester, and Nati Srebro.
\newblock Exploring generalization in deep learning.
\newblock In \emph{{Advances in Neural Information Processing Systems}}, pages
  5947--5956, 2017.

\bibitem[Nguyen et~al.(2010)Nguyen, Wainwright, and
  Jordan]{nguyen2010estimating}
XuanLong Nguyen, Martin~J Wainwright, and Michael~I Jordan.
\newblock Estimating divergence functionals and the likelihood ratio by convex
  risk minimization.
\newblock \emph{IEEE Transactions on Information Theory}, 56\penalty0
  (11):\penalty0 5847--5861, 2010.

\bibitem[Nowozin et~al.(2016)Nowozin, Cseke, and Tomioka]{nowozin2016f}
Sebastian Nowozin, Botond Cseke, and Ryota Tomioka.
\newblock f-gan: Training generative neural samplers using variational
  divergence minimization.
\newblock In \emph{Advances in Neural Information Processing Systems}, pages
  271--279, 2016.

\bibitem[Ogarrio et~al.(2016)Ogarrio, Spirtes, and Ramsey]{ogarrio2016hybrid}
Juan~Miguel Ogarrio, Peter Spirtes, and Joe Ramsey.
\newblock A hybrid causal search algorithm for latent variable models.
\newblock \emph{Conference on Probabilistic Graphical Models}, 2016.

\bibitem[Pearl(2003)]{Pearl03}
Judea Pearl.
\newblock Causality: models, reasoning, and inference.
\newblock \emph{Econometric Theory}, 19\penalty0 (675-685):\penalty0 46, 2003.

\bibitem[Pearl(2009)]{pearl2009causality}
Judea Pearl.
\newblock \emph{Causality}.
\newblock 2009.

\bibitem[Pearl and Verma(1991)]{pearl1991formal}
Judea Pearl and Thomas Verma.
\newblock \emph{A formal theory of inductive causation}.
\newblock 1991.

\bibitem[Pedregosa et~al.(2011)Pedregosa, Varoquaux, Gramfort, Michel, Thirion,
  Grisel, Blondel, Prettenhofer, Weiss, Dubourg, Vanderplas, Passos,
  Cournapeau, Brucher, Perrot, and Duchesnay]{scikitlearn}
F.~Pedregosa, G.~Varoquaux, A.~Gramfort, V.~Michel, B.~Thirion, O.~Grisel,
  M.~Blondel, P.~Prettenhofer, R.~Weiss, V.~Dubourg, J.~Vanderplas, A.~Passos,
  D.~Cournapeau, M.~Brucher, M.~Perrot, and E.~Duchesnay.
\newblock Scikit-learn: Machine learning in {P}ython.
\newblock \emph{Journal of Machine Learning Research}, 12:\penalty0 2825--2830,
  2011.

\bibitem[Peters et~al.(2014)Peters, Mooij, Janzing, and
  Sch{\"o}lkopf]{peters2014causal}
Jonas Peters, Joris~M Mooij, Dominik Janzing, and Bernhard Sch{\"o}lkopf.
\newblock Causal discovery with continuous additive noise models.
\newblock \emph{The Journal of Machine Learning Research}, 15\penalty0
  (1):\penalty0 2009--2053, 2014.

\bibitem[Peters et~al.(2017)Peters, Janzing, and Sch{\"o}lkopf]{PetJanSch17}
Jonas Peters, Dominik Janzing, and Bernhard Sch{\"o}lkopf.
\newblock \emph{Elements of Causal Inference - Foundations and Learning
  Algorithms}.
\newblock MIT Press, 2017.

\bibitem[Quinn et~al.(2011)Quinn, Mooij, Heskes, and Biehl]{quinn2011learning}
John~A Quinn, Joris~M Mooij, Tom Heskes, and Michael Biehl.
\newblock Learning of causal relations.
\newblock \emph{ESANN}, 2011.

\bibitem[Ramsey et~al.(2017)Ramsey, Glymour, Sanchez-Romero, and
  Glymour]{ramsey2017million}
Joseph Ramsey, Madelyn Glymour, Ruben Sanchez-Romero, and Clark Glymour.
\newblock A million variables and more: the fast greedy equivalence search
  algorithm for learning high-dimensional graphical causal models, with an
  application to functional magnetic resonance images.
\newblock \emph{International journal of data science and analytics},
  3\penalty0 (2):\penalty0 121--129, 2017.

\bibitem[Ramsey(2015)]{ramsey2015scaling}
Joseph~D Ramsey.
\newblock Scaling up greedy causal search for continuous variables.
\newblock \emph{arXiv}, 2015.

\bibitem[Rothenh{\"a}usler et~al.(2015)Rothenh{\"a}usler, Heinze, Peters, and
  Meinshausen]{rothenhausler2015backshift}
Dominik Rothenh{\"a}usler, Christina Heinze, Jonas Peters, and Nicolai
  Meinshausen.
\newblock Backshift: Learning causal cyclic graphs from unknown shift
  interventions.
\newblock In \emph{Advances in Neural Information Processing Systems}, pages
  1513--1521, 2015.

\bibitem[Sachs et~al.(2005)Sachs, Perez, Pe'er, Lauffenburger, and
  Nolan]{sachs2005causal}
Karen Sachs, Omar Perez, Dana Pe'er, Douglas~A Lauffenburger, and Garry~P
  Nolan.
\newblock Causal protein-signaling networks derived from multiparameter
  single-cell data.
\newblock \emph{Science}, 308\penalty0 (5721):\penalty0 523--529, 2005.

\bibitem[Schaffter et~al.(2011)Schaffter, Marbach, and
  Floreano]{schaffter2011genenetweaver}
Thomas Schaffter, Daniel Marbach, and Dario Floreano.
\newblock Genenetweaver: in silico benchmark generation and performance
  profiling of network inference methods.
\newblock \emph{Bioinformatics}, 27\penalty0 (16):\penalty0 2263--2270, 2011.

\bibitem[Scutari(2009)]{scutari2009learning}
Marco Scutari.
\newblock Learning bayesian networks with the bnlearn r package.
\newblock \emph{arXiv}, 2009.

\bibitem[Shen-Orr et~al.(2002)Shen-Orr, Milo, Mangan, and
  Alon]{shen2002network}
Shai~S Shen-Orr, Ron Milo, Shmoolik Mangan, and Uri Alon.
\newblock Network motifs in the transcriptional regulation network of
  escherichia coli.
\newblock \emph{Nature genetics}, 31\penalty0 (1):\penalty0 64, 2002.

\bibitem[Shimizu et~al.(2006)Shimizu, Hoyer, Hyv{\"a}rinen, and
  Kerminen]{shimizu2006linear}
Shohei Shimizu, Patrik~O Hoyer, Aapo Hyv{\"a}rinen, and Antti Kerminen.
\newblock A linear non-gaussian acyclic model for causal discovery.
\newblock \emph{Journal of Machine Learning Research}, 2006.

\bibitem[Spirtes and Zhang(2016)]{spirtes2016causal}
Peter Spirtes and Kun Zhang.
\newblock Causal discovery and inference: concepts and recent methodological
  advances.
\newblock \emph{Applied informatics}, 2016.

\bibitem[Spirtes et~al.(1993)Spirtes, Glymour, and Scheines]{spirtes1993search}
Peter Spirtes, Clark Glymour, and Richard Scheines.
\newblock Causation, prediction and search. 1993.
\newblock \emph{Lecture Notes in Statistics}, 1993.

\bibitem[Spirtes et~al.(2000)Spirtes, Glymour, and
  Scheines]{spirtes2000causation}
Peter Spirtes, Clark~N Glymour, and Richard Scheines.
\newblock \emph{Causation, prediction, and search}.
\newblock 2000.

\bibitem[Srivastava et~al.(2014)Srivastava, Hinton, Krizhevsky, Sutskever, and
  Salakhutdinov]{srivastava2014dropout}
Nitish Srivastava, Geoffrey Hinton, Alex Krizhevsky, Ilya Sutskever, and Ruslan
  Salakhutdinov.
\newblock Dropout: a simple way to prevent neural networks from overfitting.
\newblock \emph{The Journal of Machine Learning Research}, 15\penalty0
  (1):\penalty0 1929--1958, 2014.

\bibitem[Statnikov et~al.(2012)Statnikov, Henaff, Lytkin, and
  Aliferis]{statnikov2012new}
Alexander Statnikov, Mikael Henaff, Nikita~I Lytkin, and Constantin~F Aliferis.
\newblock New methods for separating causes from effects in genomics data.
\newblock \emph{BMC genomics}, 2012.

\bibitem[Strobl et~al.(2017)Strobl, Zhang, and
  Visweswaran]{strobl2017approximate}
Eric~V Strobl, Kun Zhang, and Shyam Visweswaran.
\newblock Approximate kernel-based conditional independence tests for fast
  non-parametric causal discovery.
\newblock \emph{arXiv preprint arXiv:1702.03877}, 2017.

\bibitem[Tsamardinos et~al.(2003)Tsamardinos, Aliferis, and
  Statnikov]{tsamardinos2003time}
Ioannis Tsamardinos, Constantin~F. Aliferis, and Alexander Statnikov.
\newblock Time and sample efficient discovery of {Markov} blankets and direct
  causal relations.
\newblock In \emph{Proceedings of the ninth ACM SIGKDD international conference
  on Knowledge discovery and data mining}, pages 673--678, 2003.

\bibitem[Tsamardinos et~al.(2006)Tsamardinos, Brown, and
  Aliferis]{tsamardinos2006max}
Ioannis Tsamardinos, Laura~E Brown, and Constantin~F Aliferis.
\newblock The max-min hill-climbing bayesian network structure learning
  algorithm.
\newblock \emph{Machine learning}, 65\penalty0 (1):\penalty0 31--78, 2006.

\bibitem[Van~den Bulcke et~al.(2006)Van~den Bulcke, Van~Leemput, Naudts, van
  Remortel, Ma, Verschoren, De~Moor, and Marchal]{van2006syntren}
Tim Van~den Bulcke, Koenraad Van~Leemput, Bart Naudts, Piet van Remortel,
  Hongwu Ma, Alain Verschoren, Bart De~Moor, and Kathleen Marchal.
\newblock Syntren: a generator of synthetic gene expression data for design and
  analysis of structure learning algorithms.
\newblock \emph{BMC bioinformatics}, 7\penalty0 (1):\penalty0 43, 2006.

\bibitem[Vergara and Est{\'e}vez(2014)]{vergara2014review}
Jorge~R. Vergara and Pablo~A. Est{\'e}vez.
\newblock A review of feature selection methods based on mutual information.
\newblock \emph{Neural computing and applications}, 24\penalty0 (1):\penalty0
  175--186, 2014.

\bibitem[Wang and Blei(2018)]{BleiBlessing}
Yixin Wang and David~M. Blei.
\newblock The blessings of multiple causes.
\newblock \emph{CoRR}, abs/1805.06826, 2018.
\newblock URL \url{http://arxiv.org/abs/1805.06826}.

\bibitem[Wang and Blei(2021)]{WangB21}
Yixin Wang and David~M. Blei.
\newblock A proxy variable view of shared confounding.
\newblock In Marina Meila and Tong Zhang, editors, \emph{Proceedings of the
  38th International Conference on Machine Learning, {ICML} 2021, 18-24 July
  2021, Virtual Event}, volume 139 of \emph{Proceedings of Machine Learning
  Research}, pages 10697--10707. {PMLR}, 2021.

\bibitem[Yu et~al.(2018)Yu, Liu, and Li]{yu2018unified}
Kui Yu, Lin Liu, and Jiuyong Li.
\newblock A unified view of causal and non-causal feature selection.
\newblock \emph{arXiv preprint arXiv:1802.05844}, 2018.

\bibitem[Zhang and Hyv{\"a}rinen(2010)]{zhang2010distinguishing}
Kun Zhang and Aapo Hyv{\"a}rinen.
\newblock Distinguishing causes from effects using nonlinear acyclic causal
  models.
\newblock In \emph{Causality: Objectives and Assessment}, pages 157--164, 2010.

\bibitem[Zhang et~al.(2012)Zhang, Peters, Janzing, and
  Sch{\"o}lkopf]{zhang2012kernel}
Kun Zhang, Jonas Peters, Dominik Janzing, and Bernhard Sch{\"o}lkopf.
\newblock Kernel-based conditional independence test and application in causal
  discovery.
\newblock \emph{arXiv}, 2012.

\bibitem[Zheng et~al.(2018)Zheng, Aragam, Ravikumar, and Xing]{Dag-no-tears}
Xun Zheng, Bryon Aragam, Pradeep Ravikumar, and Eric~P. Xing.
\newblock Dags with {NO} {TEARS:} continuous optimization for structure
  learning.
\newblock In \emph{{Advances in Neural Information Processing Systems
  (NeurIPS)}}, pages 9492--9503, 2018.

\end{thebibliography}
\end{document}